\documentclass[conference,a4paper]{IEEEtran}

\usepackage[utf8]{inputenc} %
\usepackage[T1]{fontenc}    %
\usepackage{hyperref}       %
\usepackage{url}            %
\usepackage{booktabs}       %
\usepackage{amsfonts}       %
\usepackage{nicefrac}       %
\usepackage{microtype}      %

\usepackage[]{graphicx}
\usepackage{caption}
\usepackage{subfig}
\usepackage{amssymb}
\usepackage{amsmath}
\usepackage{gensymb}
\usepackage{xcolor}
\usepackage{epsfig} %
\usepackage{balance}
\usepackage{array}
\usepackage{siunitx}
\usepackage{ragged2e}
\usepackage{comment}
\usepackage{hyperref}
\hypersetup{
    colorlinks=true,
    linkcolor=blue,
    filecolor=magenta,      
    urlcolor=cyan,
}

\usepackage[switch]{lineno}

\usepackage{tabularx}

\newcommand{\etal}{\emph{et~al.}}

\ifCLASSINFOpdf
\else
\fi
\IEEEoverridecommandlockouts
\begin{document}
\title{\LARGE \bf{Multi-species Seagrass Detection and Classification \\ from Underwater Images}}
\author{Scarlett Raine$^{1,2}$, Ross Marchant$^{1,2}$, Peyman Moghadam$^{1}$, Frederic Maire$^{2}$, Brett Kettle$^{1}$, Brano Kusy$^{1}$
\\
$^{1}$ CSIRO Data61, Brisbane, Australia
\\
$^{2}$ School of Electrical Engineering and Robotics, Queensland University of Technology (QUT), Brisbane, Australia \\
\thanks{$^{1}$ E-mails: {\tt\footnotesize \emph{scarlett.raine, ross.marchant, peyman.moghadam, brett.kettle, brano.kusy}@csiro.au}}
\thanks{$^{2}$ E-mails: {\tt\footnotesize \emph{r.marchant, f.maire@qut.edu.au}}
} 
}

\maketitle
\thispagestyle{plain}
\pagestyle{plain}

\IEEEpeerreviewmaketitle
\begin{abstract}
Underwater surveys conducted using divers or robots
equipped with customized camera payloads can generate a large number of images. 
Manual review of these images to extract ecological data is prohibitive in terms of time and cost, 
thus providing strong incentive to automate this process using machine learning solutions. 
In this paper, we introduce a multi-species detector and classifier for seagrasses based on a deep convolutional neural network (achieved an overall accuracy of 92.4\%). We also introduce a simple method to semi-automatically label image patches and therefore minimize manual labelling requirement. We describe and release publicly the dataset collected in this study as well as the code and pre-trained models to replicate our experiments at: \href{https://github.com/csiro-robotics/deepseagrass}{https://github.com/csiro-robotics/deepseagrass}

\end{abstract}
\section{Introduction}
\label{sec:intro}  
Seagrasses are high value coastal resources that provide a range of ecosystem services, including physical protection of coastlines, mitigation of adverse water quality, increasing fisheries productivity, and acting as `blue carbon' sinks for mitigating climate change \cite{lavery}.  Active management of these systems requires assessment of the extent and condition of seagrass meadows, a complex task because they are dynamic in space and time, both naturally and in response to anthropogenic impacts \cite{unsworth}. Well-studied meadows are in decline globally \cite{unsworth2}, highlighting the need to manage large areas of extant meadows, as well as their historical and potential habitats. An updated assessment of known distributions concludes that 166,387 \SI{}{\km\squared} of meadows (moderate to high probability assessment) currently exist \cite{vanderklift}. Recent modelling, based on depth, minimum photosynthetic requirements and seagrass distributional studies, suggest a potential global seagrass habitat of 1,646,788 \SI{}{\km\squared}. In part, the ten-fold difference between known and potential distributions relates to under-survey or limitations of existing survey methods \cite{mckenzie2020} in environments where surveys are impeded by remoteness, turbidity, depth and coastal conditions \cite{mckenzie2001}.

Recognising the importance of seagrass and the necessity of its monitoring for better management, there have been a number of advances in vision-based seagrass survey methods to increase the spatial extent or intensity of surveys beyond that which can be accomplished on foot at low tide, or by snorkelers or divers. Broad-scale methods include remote sensing (satellites, conventional aerial photography, drones) and underwater vehicles that are towed (TUVs \cite{rende}), remotely operated (ROVs \cite{finkl}) or fully autonomous (AUVs \cite{monk}). Computer vision and machine learning techniques are used to automatically and efficiently analyse the large amount of image data produced \cite{coralnet}. 
Real-time processing of images is essential if monitoring utilises an underwater robot and the objective requires adaptation of sampling effort or survey extents during the field program.

In this paper we present two novel contributions to the automatic seagrass detection and classification problem. Firstly, we have created a new publicly available image dataset (called \textit{DeepSeagrass}) of three seagrass morphotypes from Moreton Bay, Australia. The dataset is unique in that it contains high-resolution images at an oblique angle and close range, which closely matches the perspective of a diver or underwater robot, and consists primarily of images with only a single morphotype present.  This  allows image-level labels to be assigned to subregions within the images.
Secondly, we designed a deep neural network based seagrass classification system and evaluated it on this image dataset.  The training of the neural network takes advantage of the single-morphotype images to learn to identify seagrass in small image patches. The advantages of a patch-based approach are its simplicity, easy application to real-time video from an underwater robot, and its ability to yield spatial data at a sub-image level to facilitate estimates of percentage cover or mapping outputs. 

After reviewing existing work on seagrass monitoring in Section \ref{sec:related}, we describe the dataset that we release publicly in Section \ref{sec:datasets}. In Section \ref{sec:method}, we explain the semi-automatic labeling of the dataset,  the architecture of the neural classifier and details of the training process. Experimental results are reported in Section \ref{sec:results}.

\section{Related Work}
\label{sec:related}

Improving the mapping of seagrass beds is an active area of research within the marine biology community. Seagrass data is collected using techniques such as manual surveys \cite{roelfsema}, helicopter surveys \cite{campbell}, remote sensing such as satellite imagery \cite{phinn} or AUVs \cite{bonin-font}, \cite{burgeura}, \cite{martin-abadal}; or a combination of two or more methods \cite{dunbabin}. Producing seagrass maps from the vast amounts of data is often time-consuming using manual counting methods, and thus automated methods are an active research area.

\subsection{Seagrass Identification}

Two traditional machine learning approaches to seagrass detection use Gabor filtering \cite{bonin-font}, \cite{burgeura} or gray-level co-occurrence matrices \cite{massot-campos} to generate feature descriptors, which are then used to train a Support Vector Machine (SVM) for pixel or patch level classification. More recently, deep learning using convolutional neural networks (CNNs) has emerged as the state of the art in classification problems \cite{Sultana_2018}, and thus has potential for the automated classification of seagrass, including accurate real-time calculation of percentage cover \cite{monir_review}. 

Moniruzzaman \etal~\cite{moniruzzaman} used the Faster-RCNN \cite{faster-rcnn} network to detect individual leaves of a single species of seagrass, \textit{Halophila ovalis}, however, they did not consider multiple species or where the seagrass appears as a dense carpet over the seabed. Reus \etal~\cite{reus} achieved 94.5\% accuracy on pixel-wise segmentation of seagrass, later improved on by Weidmann \etal~\cite{weidmann}, in images taken by an AUV vertically downward. They used rectangular image patches to train a CNN to produce binary pixel-wise classifications, but did not consider different species of seagrass. Likewise, Martin-Abadal \etal~ \cite{martin-abadal} compared a CNN with other methods of machine learning for single species, finding that a VGG-16 encoder with an 8-stride fully convolutional network as the decoder \cite{Long2015} achieved state-of-the-art accuracy for pixel-wise segmentation of the seagrass \textit{Posidonia oceanica}.

In this work, we address the problem of identifying multiple species of seagrass. Our approach is to classify different morphotypes at the image patch level, using a CNN trained in a fully-supervised manner. However, due to the cost of labelling at a patch or even pixel level, we use images that have only one type of seagrass so that the image-level labels can be applied as weak labels to the patches.

\begin{figure}[t]
\centering
  \subfloat[][Strappy] {
    \centering
    \includegraphics[width=0.20\textwidth]{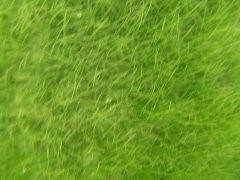}}
  \subfloat[][Ferny] {
    \centering
    \includegraphics[width=0.20\textwidth]{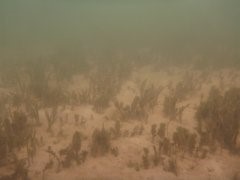}}\\
    \subfloat[][Rounded] {
    \centering
    \includegraphics[width=0.20\textwidth]{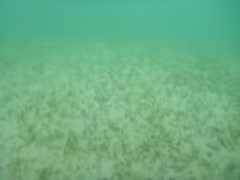}}
    \subfloat[][Background] {
    \centering
    \includegraphics[width=0.20\textwidth]{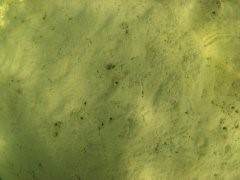}}
  \caption{Sample of different species of seagrass from a public image dataset \cite{roelfsema}.}
  \label{figure:public_images}
\end{figure}

\subsection{Existing Seagrass Datasets}

The method requires a seagrass dataset representing multiple species, with enough images containing only one particular species. A survey of existing publicly available datasets (Table \ref{table:all_datasets}) found that there does not exist, to our knowledge, a dataset designed for classification and detection of different species of seagrass.  Only two datasets contained multiple species of seagrass (entries 2 and 5 in Table \ref{table:all_datasets}) and of these, only the photo-transect series of the Eastern Banks of Moreton Bay (entry 2 in Table \ref{table:all_datasets}) is publicly available, referred to as the `Public Dataset' in the rest of this paper.
Figure \ref{figure:public_images} shows a sample of the images in this dataset \cite{roelfsema}.
`Public Dataset' images were collected on the Eastern Banks of Moreton Bay, near Brisbane, Australia and have been made available on Pangaea \cite{roelfsema}. Image resolution is very low, with sizes of 105x140, 150x200 and 180x240 pixels (Figure \ref{figure:public_images}). Images contain a variety of marine species and are accompanied at the image-level with Coral Point Count (CPC) annotations \cite{roelfsema}. The CPC method is a manual computer-assisted technique that overlays points on an image.  Labels are assigned to the element directly under a point, and cover is estimated based on point counts \cite{cpc}. 

The images were labelled into three morphological taxonomic super-classes according to their CPC label (plus one substrate class): `Strappy' encompassed \textit{Cymodocea serrulata}, \textit{Halodule uninervis}, \textit{Syringodium isoetifolium} and \textit{Zostera muelleri}; `Ferny' described \textit{Halophila spinulosa}; `Rounded' described \textit{Halophila ovalis}; and `Background' consisted of images with less than 1\% total seagrass cover.
Images with less than 70\% cover for one species were removed from the dataset. Furthermore, due to the coarse sampling of the CPC method, some images did not have accurate species estimations.  The dataset was manually inspected by a domain expert and incorrectly labeled images were removed. The filtered dataset was heavily skewed towards the Strappy class, with significantly fewer examples of Ferny and Rounded seagrass (see Table \ref{table:publicdata}).

\begin{table}[t]
\begin{center}
\caption{Class Distribution in the Public Dataset \cite{roelfsema}} \label{table:publicdata}
\begin{tabular}{ |p{1.3cm}|p{1.3cm}|p{1.3cm}|p{1.6cm}| } 
\hline
\textbf{Strappy} & \textbf{Ferny} & \textbf{Rounded} & \textbf{Background}\\ 
\hline
4671 & 76 & 34 & 1088 \\ 
\hline
\end{tabular}
\end{center}
\end{table}

\begin{table*}[h]
\caption{Existing Seagrass Datasets}
\begin{center}
\begin{tabular}{ |p{0.2cm}|p{3.5cm}|p{0.7cm}|p{1.7cm}|p{1.8cm}|p{3.5cm}|p{1.7cm}| }
    \hline
    \textbf{ } & \textbf{Name of Paper} & \textbf{Year} & \textbf{Dataset Size} & \textbf{Annotations} 
    & \textbf{Seagrass Species} & \textbf{Access}\\ \hline
    1 & \raggedright Faster R-CNN Based Deep Learning for Seagrass Detection from Underwater Digital Images \cite{moniruzzaman} & 2019 & \raggedright 2,699 images of Halophila ovalis & Bounding boxes & Halophila ovalis & No public access \tabularnewline \hline 
    2 & \raggedright Benthic and substrate cover data derived from a time series of photo-transect surveys for the Eastern Banks, Moreton Bay Australia \cite{roelfsema} & 2004-2015 & \raggedright 8 datasets, each with approximately 3,000 images & \raggedright Seagrass species and percentage cover for each frame generated using Coral Point Count & \raggedright 41 subcategories, including Halophila ovalis, Halophila spinulosa, Cymodocea serrulata, Zostera muelleri and Syringodium isoetifolium, and other biota such as algae, starfish, sea cucumbers   & \raggedright Publicly available on Pangaea \tabularnewline \hline
    3 & \raggedright Looking for Seagrass: Deep Learning for Visual Coverage Estimation \cite{reus} & 2018 & \raggedright 12682 total images, 6037 annotated & Binary polygons indicating if a pixel belongs to the class seagrass or background & Unspecified & \raggedright Images and annotations publicly available on Github \tabularnewline \hline
    4 & \raggedright IMOS - AUV SIRIUS Great Barrier Reef Campaign \cite{IMOS}, \cite{bewley2015australian} & 2011 & \raggedright 9,874 images, but only 44 seagrass annotation points & Labelled using Coral Point Count as described in \cite{bewley2015australian} & Points broadly labelled as `seagrass' & \raggedright Publicly available on the IMOS website \tabularnewline \hline
    5 & \raggedright Seagrass meadows of Hervey Bay and the Great Sandy Strait, Queensland, derived from field surveys conducted 6-14 December, 1998 \cite{mckenzie} & 2017 & \raggedright 1,104 field validation points - videos were recorded for points greater than 10m depth & \raggedright Seagrass composition recorded according to traditional survey methods & \raggedright Cymodocea serrulata, Halodule uninervis, Syringodium isoetifolium, Halophila decipiens, Halophila ovalis, Halophila spinulosa and Zostera muelleri & \raggedright Survey data publicly available on Pangaea, but no access to images or video  \tabularnewline \hline
\end{tabular}
\end{center}
\label{table:all_datasets}
\end{table*}

\section{DeepSeagrass Dataset} 
\label{sec:datasets}
Due to the poor resolution, class imbalance, small image sizes, inconsistent camera angles, high level of blur and the overall small quantity of usable images, we decided to collect a new dataset from a similar location. 

Images were acquired across nine different seagrass beds in Moreton Bay, Australia over four days during February 2020. Search locations were chosen according to distributions reported in the publicly available dataset. A biologist made a search of each area, snorkelling in approximately 1 - 2m of water during low to mid tide. In-situ search of seagrass beds resulted in batches of photographs in 78 distinct geographic sub-areas, each containing one particular seagrass morphotype (or bare substrate).  Images were taken using a Sony Action Cam FDR-3000X from approximately 0.5m off the seafloor at an oblique angle of around 45 degrees. Over 12,000 high-resolution (4624 x 2600 pixels) images were obtained. 

Images were reviewed to ensure that any containing a second morphotype at more than approximately 0.5\% density were placed into a `mixed' class. The remainder were labelled according to their dominant morphotype, and divided into three categories (dense, medium, sparse) according to the relative density of seagrass present, for a total of 11 classes.

\begin{figure}
    \centering
    \includegraphics[width=0.45\textwidth]{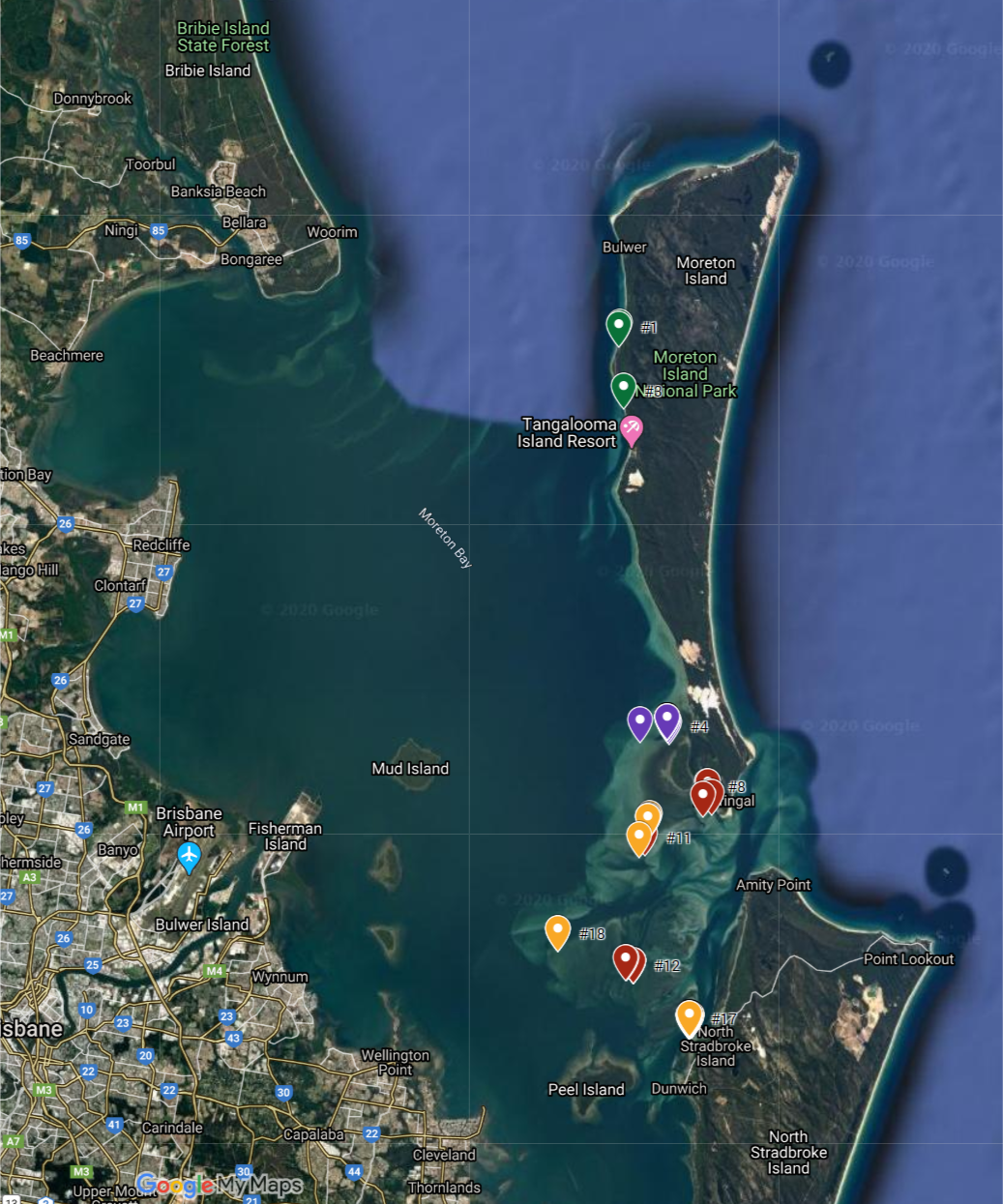}
    \caption{Map of 78 distinct sub-areas for our \textit{DeepSeagrass} dataset. Colour of the marker refers to the date of collection, green: 2020-02-07, purple: 2020-02-08, red: 2020-02-10, orange: 2020-02-11.}
    \label{fig:map}
\end{figure}

\indent The dataset was then prepared for use in a machine learning pipeline by taking the dense seagrass images and dividing them into train and test sets. Images from 70 of the sub-areas were allocated to the training set (1,701 images in total), while images from the remaining eight sub-areas were reserved exclusively for the test set (335 images). Different sub-areas were used so that evaluation of a classification system on the test set would assess how well it generalised to different geographic areas.

\section{Method}
\label{sec:method}
Our approach classifies seagrass on a per-patch basis using a pre-trained CNN classifier. A patch-based method was used because it captures location information about the seagrass in the frame without the need for dense polygon, pixel or bounding box labels within each training image.

\subsection{Dataset Preparation}
 We use underwater images that consist of only one seagrass morphotype in each image. For the \textit{DeepSeagrass} dataset, these are the `Ferny - dense', `Strappy - dense' and `Round - dense' classes, plus images with no seagrass from the `Substrate' class.
 
\begin{figure}[t]
\centering
\includegraphics[scale=0.4]{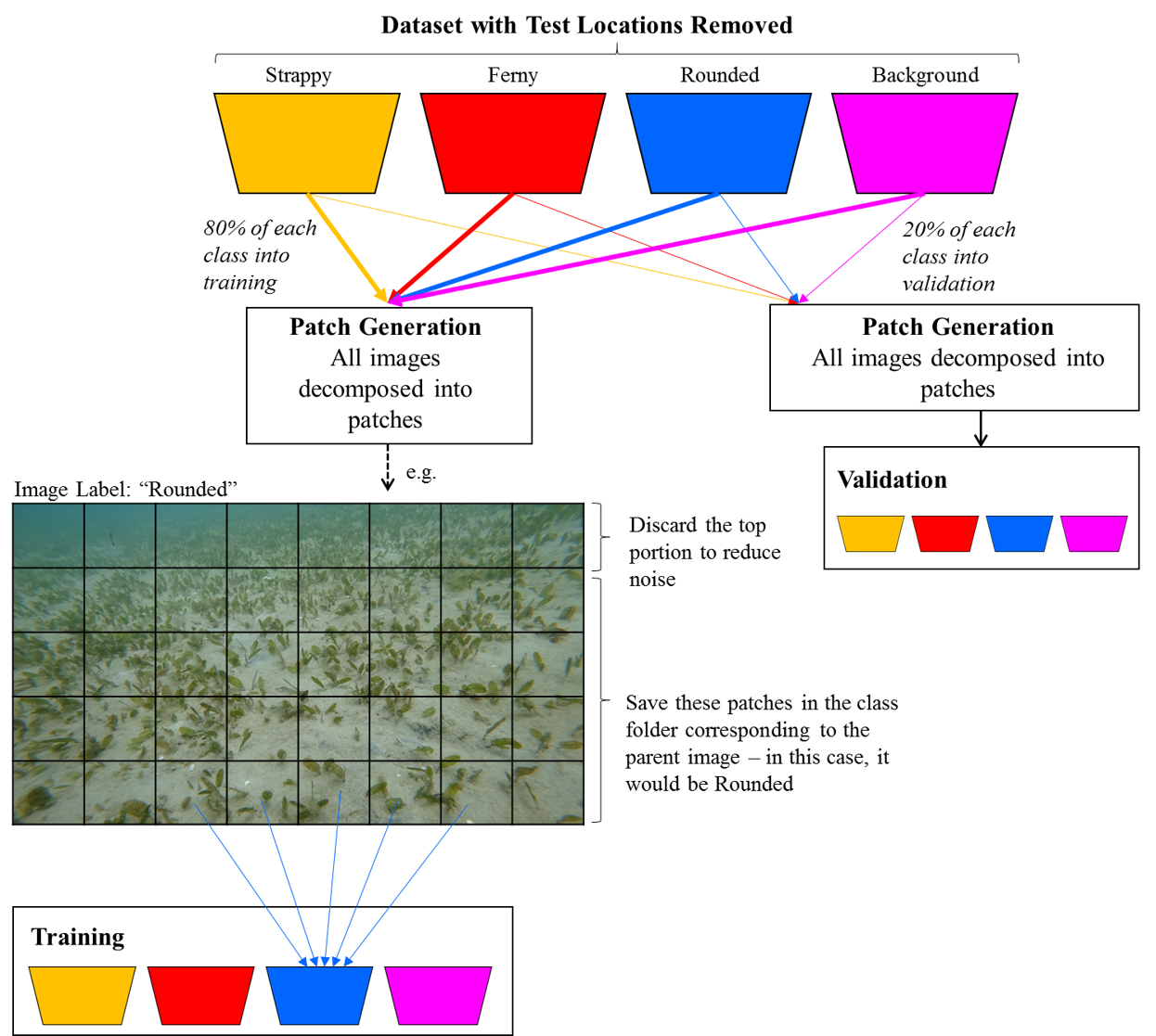}
\caption{Overview of our dataset preparation pipeline.}
\label{fig:data_gen}
\end{figure}

 Images are then divided into a grid of patches. For the \textit{DeepSeagrass} dataset we used a grid of 5 rows by 8 columns to generate 40 patches of size 520x578 pixels per image (as discussed in Section \ref{subsubsec:patch_size}). The top row of patches for each image was discarded, as their oblique pose in low underwater visibility conditions frequently resulted in hard to distinguish seagrass (Figure \ref{fig:data_gen}).

 Based on our data collection procedure, we can assume (weak labels) that the remainder of the patches for an image represent the same seagrass species (or are all substrate). As such, each patch is assigned the image label, reducing the time and cost of labelling considerably. Patches from the training set were further split into 80\% training and 20\% validation sets. Together with the test set patches, these form the dataset of approximately 66,946 image patches (Table \ref{table:testdataset}) that are used for all experiments in this paper.

\begin{table}[t]
\caption{
Class Distribution of Image Patches in \emph{DeepSeagrass} Dataset
}
\begin{center}
\begin{tabular}{ |c|c|c|c|c| }
    \hline 
        \multicolumn{5}{|c|}{\textbf{Training}}\\
    \hline
    \textbf{Strappy} & \textbf{Ferny} & \textbf{Rounded} & \textbf{Background} & \textbf{Total} \\ 
    \hline
    11,584 & 8,256 & 13,792 & 9,216 & 42,848 \\ 
    \hline
    \multicolumn{5}{|c|}{\textbf{Validation}}\\
    \hline
    \textbf{Strappy} & \textbf{Ferny} & \textbf{Rounded} & \textbf{Background} & \textbf{Total}\\ 
    \hline
    2,880 & 2,080 & 3,456 & 2,304 & 10,720 \\
    \hline 
        \multicolumn{5}{|c|}{\textbf{Test}}\\
    \hline
    \textbf{Strappy} & \textbf{Ferny} & \textbf{Rounded} & \textbf{Background} & \textbf{Total} \\ 
    \hline
    2,643 & 4,447 & 1,345 & 4,943 & 13,378 \\
    \hline 
\end{tabular}
\end{center}
\label{table:testdataset}
\end{table}

\subsection{Classification Model}
We use a CNN based on the VGG-16 architecture \cite{Simonyan15} for the image classifier. It consists of the VGG-16 model pretrained on the ImageNet classification task, with final dense layers removed and replaced with the following consecutive layers: dense (with 512 nodes and ReLU \cite{agarap} activation), dropout \cite{srivastava} (with dropout probability of 0.05), dense (with 512 nodes and ReLU activation), dropout (with dropout probability of 0.15) and a final dense layer with one node for each class, activated with the Softmax function. Using heavy dropout in the final layers has been successful for other transfer learning approaches with small numbers of classes \cite{srivastava}.

\subsection{Training}
The CNN was trained on batches of 32 images at each step, using the Adam optimiser \cite{kingma} with learning rate set to an initial value of 0.001. The learning rate was halved whenever there was no 
improvement in loss over the last 10 epochs, and after four such decreases the training was stopped.

Colour-based image augmentation was applied to the training images to make the model robust to underwater light variations driven by the incident angle of the sun striking the surface of the water, the turbidity and colour of the water and the depth at which imagery is captured. To mimic these effects, we chose to vary the brightness, contrast and blur for data augmentation.  Additionally, the red channel was altered to change the "blueness" of the images. The data augmentation ablation study is described in Section \ref{subsubsec:data_aug}.

\subsection{Application to Whole Images}

\indent The trained model is then applied to whole frames from an underwater video.  Whole images are divided into patches with size equal to that used to train the model.  Each patch is classified, assigning a label corresponding to the argmax of the Softmax output layer. 
Multi-label classification is also possible by 
replacing the Softmax layer with an array of Sigmoid units,
but we do not cover this approach in this paper. 
Finally, classification labels are visualised by colour coded overlay onto the original image.

\section{Results}
\label{sec:results}

The classifier was trained and evaluated using 5-fold cross validation on the training set, whereby the training set is split into 80\% training and 20\% validation sets, and the evaluation metrics were averaged across all five runs. The metrics used for evaluating the performance of the models were per-class precision, recall and the overall accuracy. The precision and recall were calculated for each of the four classes.  Additionally, an overall accuracy was calculated by determining the fraction of correctly classified examples over all images. 
The model achieved 98.2\% to 99.6\% precision and 98.0\% to 99.7\% recall for each class, and an overall accuracy of 98.8\%. Accuracy dropped to 88.2\% when the model was applied to test patches from the unseen geographic sub-areas, suggesting that images from these sites are different (Tables \ref{table:data_aug_val_results} and \ref{table:data_aug_test_results}).

\subsection{Ablation Study}
\label{subsec:ablation}

An ablation study was performed to see the effect of model architecture, patch size and augmentation methods. All experiments not involving the test set were implemented using 5-fold cross validation.

\subsubsection{Network Architecture}

In the first experiment we tested two different feature extractors, the VGG-16 \cite{Simonyan15} and ResNet-50 \cite{he2015deep} CNNs pretrained on ImageNet with final dense layers removed. One of three final layer configurations on the output of these networks were used: first, k-Nearest Neighbours (``KNN'') with k=3. Second, a simple 2-layer dense network (``2-Layer'') with 16 nodes and 14 nodes consecutively. Last, a 2-layer dense network with dropout of 0.05 between each dense layer of 512 nodes (``2-Layer+Drop'').
Initial experiments showed no real difference between any configuration, so instead the models were evaluated on the test set. The VGG-16 CNN with ``2-Layer+Drop'' configuration was the best performing network with 87.2\% overall accuracy. Interestingly, the overall accuracy of ResNet-50 network did not change with the final configuration (Table \ref{table:test_results}).

\newcolumntype{a}{>{\hsize=0.7\hsize}X}
\newcolumntype{b}{>{\hsize=3.7\hsize}X}
\begin{table*}[h]
\centering
\caption{Performance of Classifiers on the Unseen Test Dataset}
\label{table:test_results}
\begin{tabularx}{\textwidth}{baaaaaaaaa}
\toprule
{} &
\multicolumn{2}{c}{Strappy} &
\multicolumn{2}{c}{Ferny} & 
\multicolumn{2}{c}{Rounded} & 
\multicolumn{2}{c}{Background} & {}\\
\midrule
\textbf{Model}   & \text{Prec.} & \text{Recall} & \text{Prec.} & \text{Recall} & \text{Prec.} & \text{Recall} & \text{Prec.} & \text{Recall} & \text{Acc.} \\
\midrule
VGG-16 KNN   & 0.76 & 0.84 & 0.80 & 0.91 & 0.73 & 0.83 & 0.97 & 0.75 & 0.83\\
VGG-16 2-Layer  & 0.80 & 0.96 & 0.82 & 0.94 & \textbf{0.89} & 0.84 & \textbf{0.98} & 0.76 & \textbf{0.87} \\
VGG-16 2-Layer+Drop. & \textbf{0.82} & 0.97 & 0.83 & \textbf{0.95} & 0.86 & 0.84 & \textbf{0.98} & 0.76 & \textbf{0.87} \\
ResNet-50 KNN   & \textbf{0.82} & 0.87 & 0.83 & \textbf{0.95} & 0.67 & 0.83 & 0.97 & 0.75 & 0.85 \\
ResNet-50 2-Layer   & 0.65 & \textbf{0.99} & \textbf{0.90} & 0.85 & 0.88 & \textbf{0.87} & \textbf{0.98} & 0.76 & 0.85 \\
ResNet-50 2-Layer+Drop.   & 0.74 & 0.84 & 0.83 & 0.94 & 0.83 & 0.76 & 0.97 & \textbf{0.78} & 0.85 \\
\bottomrule
\end{tabularx}
\end{table*}

\subsubsection{Patch Size} 
\label{subsubsec:patch_size}

\begin{figure}[t]
\centering
  \subfloat[][260x289 Pixels] {
    \centering
    \includegraphics[width=0.4\textwidth]{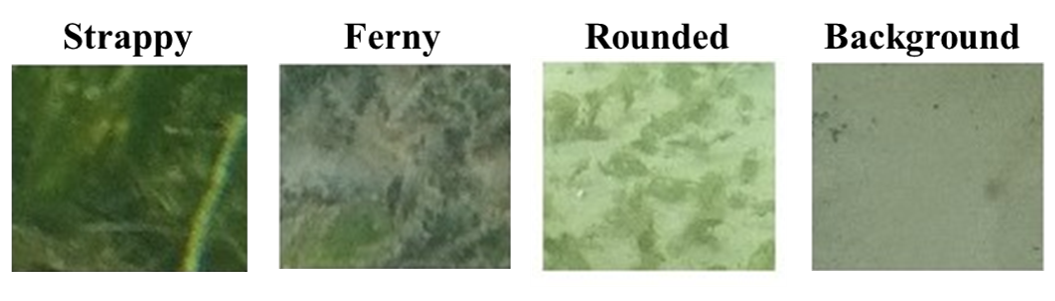}} \\
  \subfloat[][520x578 Pixels] {
    \centering
    \includegraphics[width=0.4\textwidth]{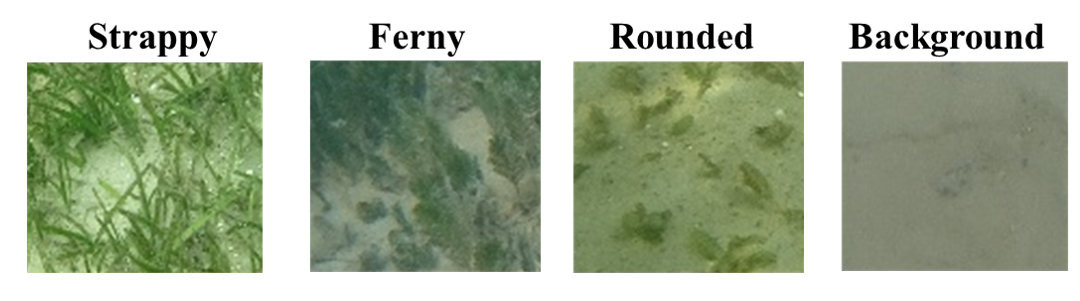}}
  \caption{Different patch size during the training phase.}
  \label{fig:patch_size}
\end{figure}

The best performing network of the previous section (VGG-16 with 2-Layer+Drop) was used to investigate the effect of patch size and data augmentation. The patch size was considered as a hyper-parameter of the training phase. Smaller patch size did not improve model performance on the validation set, but decreased per-class precision and recall (Table \ref{table:patch_size_results}). This may be because the smaller patch size removed discriminative features.  The effect was emphasised for the Strappy and Ferny morphotypes, which generally appear larger in the image, while the smaller patch size was more suitable for physically smaller Rounded morphotypes (Figure \ref{fig:patch_size}).

\begin{table*}[h]
\centering
\newcolumntype{a}{>{\hsize=0.7\hsize}X}
\newcolumntype{b}{>{\hsize=3.7\hsize}X}
\caption{Effect of Patch Size on Per-Class Precision and Recall}
\label{table:patch_size_results}
\begin{tabularx}{\textwidth}{baaaaaaaaa}
\toprule
{} &
\multicolumn{2}{c}{Strappy} &
\multicolumn{2}{c}{Ferny} & 
\multicolumn{2}{c}{Rounded} & 
\multicolumn{2}{c}{Background} & {}\\
\midrule
\textbf{Patch Size}   & \text{Prec.} & \text{Recall} & \text{Prec.} & \text{Recall} & \text{Prec.} & \text{Recall} & \text{Prec.} & \text{Recall} & \text{Acc.} \\
\midrule
VGG-16 520x578 & \textbf{1.0000} & \textbf{0.9639} & \textbf{0.9655} & \textbf{1.0000} & \textbf{0.9826} & \textbf{0.9496} & 0.9242 & \textbf{0.9839} & \textbf{0.9688} \\
VGG-16 260x289 &   0.9501 & 0.9195 & 0.9269 & 0.9234 & 0.9475 & 0.9193 & \textbf{0.9511} & 0.9822 & 0.9395 \\
\midrule
ResNet-50 520x578 &  \textbf{0.9873} & \textbf{0.9398} & \textbf{0.9667} & \textbf{1.0000} & \textbf{0.9725} & \textbf{0.9725} & \textbf{0.9857} & 0.9857 & \textbf{0.9781} \\
ResNet-50 260x289 &  0.9547 & 0.9358 & 0.9383 & 0.9570 & 0.9593 & 0.9278 & 0.9712 & \textbf{0.9912} & 0.9541 \\
\bottomrule
\end{tabularx}
\end{table*}

\subsubsection{Data Augmentation} 
\label{subsubsec:data_aug}
Five data augmentation approaches were tested: no data augmentation; horizontal flipping 50\% of the time; geometric augmentation, consisting of horizontal flips, random cropping, zooming and translation; colour augmentation, consisting of contrast, brightness, Gaussian blur and alteration of the red channel; and a combination of both the geometric and colour augmentations.

\begin{table*}[h]
\centering
\newcolumntype{a}{>{\hsize=0.7\hsize}X}
\newcolumntype{b}{>{\hsize=3.7\hsize}X}
\caption{Effect of Data Augmentation on 5-Fold Cross Validation Per-Class Precision and Recall}
\label{table:data_aug_val_results}
\begin{tabularx}{\textwidth}{baaaaaaaaa}
\toprule
{} &
\multicolumn{2}{c}{Strappy} &
\multicolumn{2}{c}{Ferny} & 
\multicolumn{2}{c}{Rounded} & 
\multicolumn{2}{c}{Background} & {}\\
\midrule
\textbf{Augmentation Type}   & \text{Prec.} & \text{Recall} & \text{Prec.} & \text{Recall} & \text{Prec.} & \text{Recall} & \text{Prec.} & \text{Recall} & \text{Acc.} \\
\midrule
No Data Aug. & \text{0.993} & \textbf{0.994} & 0.991 & 0.983 & \textbf{0.993} & 0.993 & 0.983 & 	0.992 & \textbf{0.991} \\
Horizontal Flipping Only & \textbf{1.000} & 0.980 & 0.990 & 0.982 & 0.987 & \textbf{0.993} & 0.985 & 0.988 & 0.989 \\
Geometric Aug. & 0.995 & 0.976 & \textbf{0.997} & \textbf{0.990} & 0.986 & 0.993 & \textbf{0.986} & 0.995 & 0.990 \\
Colour Aug. & 0.990 & 0.980 & 0.996 & 0.982 & 0.983 & 0.990 & 0.982 & 0.997 & 0.988 \\
Geometric \& Colour Aug. & 0.993 & 0.966 & 0.992 & 0.968 & 0.977 & 0.961 & 0.969 & \textbf{1.000} & 0.982 \\
\bottomrule
\end{tabularx}
\end{table*}

\begin{table*}[!h]
\centering
\caption{Effect of Data Augmentation on Test Dataset Per-Class Precision and Recall}
\label{table:data_aug_test_results}
\begin{tabularx}{\textwidth}{baaaaaaaaa}
\toprule
{} &
\multicolumn{2}{c}{Strappy} &
\multicolumn{2}{c}{Ferny} & 
\multicolumn{2}{c}{Rounded} & 
\multicolumn{2}{c}{Background} & {}\\
\midrule
\textbf{Augmentation Type}   & \text{Prec.} & \text{Recall} & \text{Prec.} & \text{Recall} & \text{Prec.} & \text{Recall} & \text{Prec.} & \text{Recall} & \text{Acc.} \\
\midrule
No Data Aug. & 0.82 & \textbf{0.966} & 0.828 & 0.95 & 0.86 & 0.844 & \textbf{0.98} & 0.758 & 0.872 \\
Horizontal Flipping Only & 0.83 & 0.954 & 0.822 & 0.954 & 0.858 & 0.848 & \textbf{0.98} & 0.76 & 0.87 \\
Geometric Aug. & 0.848 & 0.956 & 0.808 & 0.96 & \textbf{0.864} & 0.836 & \textbf{0.98} & 0.752 & 0.87 \\
Colour Aug. & 0.836 & 0.954 & \textbf{0.842} & \textbf{0.964} & 0.862 & \textbf{0.86} & 0.976 & \textbf{0.774} & \textbf{0.882} \\
Geometric \& Colour Aug. & \textbf{0.854} & 0.946 & 0.822 & 0.962 & 0.85 & 0.856 & 0.976 & 0.774 & 0.878 \\
\bottomrule
\end{tabularx}
\end{table*}

For the data augmentation experiments, the results from 5-fold cross validation showed very little difference between augmentation types (Table \ref{table:data_aug_val_results}).  Evaluation on the test dataset showed more variation (Table \ref{table:data_aug_test_results}). The best performing strategy was colour augmentation, followed by colour and geometric augmentation. This suggests that the lighting, turbidity or depth of the test dataset may have been outside the distribution of the training dataset, and that colour augmentation is important for robust seagrass classification.

\subsection{t-SNE Analysis}
\begin{figure}[b]
\centering
  \subfloat[][Training Dataset] {
    \centering
    \includegraphics[width=0.20\textwidth]{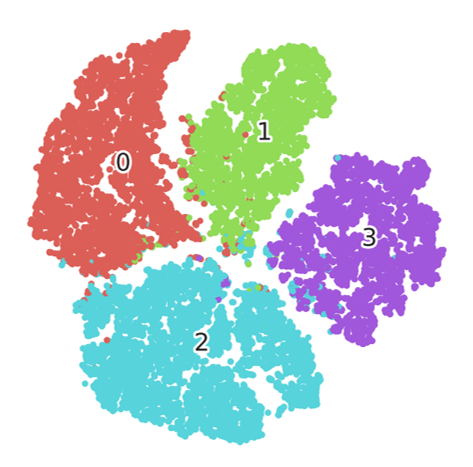}}
  \subfloat[][Test Dataset] {
    \centering
    \includegraphics[width=0.20\textwidth]{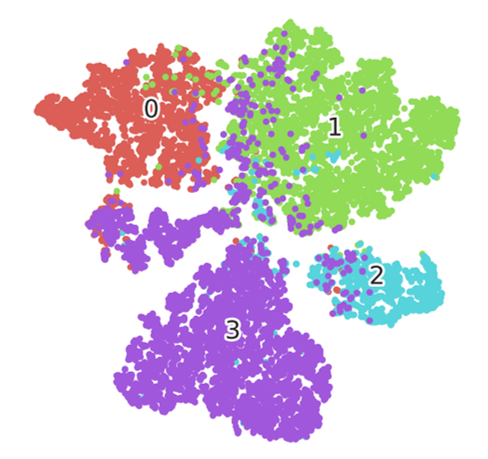}}
  \caption{t-SNE plots from training and test datasets.}
  \label{fig:tsne_1}
\end{figure}

The output of the penultimate layer in the classification model (before the Softmax output layer) was considered to be a feature vector representing the input image. For visualisation, dimensionality reduction was performed using t-SNE~\cite{tsne} on the feature vectors extracted from both the training and test datasets (Figure \ref{fig:tsne_1}). 
The resulting t-SNE plots were analysed to see if the clusters in the low-dimensional representation correspond to the image classes, as this provides an insight into the ability of the classifier to extract useful features for class discrimination (Figure \ref{fig:tsne_1}).

The training set feature vectors are grouped into four clusters that generally correspond to their classes. There are some exceptions which may be explained by patches being assigned incorrect image-level labels, for example, an area of bare substrate in an otherwise densely covered seagrass image, as well as blurry and indistinct patches from the far field of view.

\begin{figure}[ht]
\centering
\begin{center}
\includegraphics[scale=0.4]{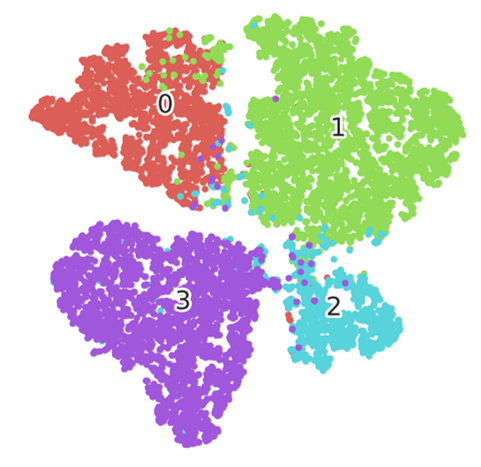}
\caption{t-SNE plots from test dataset with water images removed.} \label{figure:tsne_nowater}
\end{center}
\end{figure}

By contrast, there are two distinct clusters for class 3 ("Background") apparent in the test t-SNE plot, when only one is present in the training plot (Figure \ref{fig:tsne_1}). This can be attributed to image patches of the water column, present in the test dataset but not in the training dataset.  Water column patches were not separated into the Background class during training dataset preparation as all child patches from each seagrass image were assigned to that class.  When water column patches are removed from the test dataset, the extra cluster in the t-SNE plot disappeared (Figure \ref{figure:tsne_nowater}) and the test accuracy of the classifier improved (Table \ref{table:no_water}).  

\begin{table*}[h]
\centering
\newcolumntype{a}{>{\hsize=0.7\hsize}X}
\newcolumntype{b}{>{\hsize=3.7\hsize}X}
\caption{4-Class Classifier Per-Class Precision and Recall with Water Patches Removed from Test Dataset}
\label{table:no_water}
\begin{tabularx}{\textwidth}{baaaaaaaaa}
\toprule
{} &
\multicolumn{2}{c}{\textbf{Strappy}} &
\multicolumn{2}{c}{\textbf{Ferny}} & 
\multicolumn{2}{c}{\textbf{Rounded}} & 
\multicolumn{2}{c}{\textbf{Substrate}} & \\
\midrule
{} & \text{Prec.} & \text{Recall} & \text{Prec.} & \text{Recall} & \text{Prec.} & \text{Recall} & \text{Prec.} & \text{Recall} & \text{Accuracy} \\
\midrule
Validation Results & 0.990 & 1.000 & 0.982 & 0.985 & 0.990 & 0.958 & 0.991 & 1.000 & 0.988 \\
Test Results & 0.96 & 0.96 & 0.97 & 0.97 & 0.88 & 0.88 & 0.97 & 0.98 & 0.96  \\
\bottomrule
\end{tabularx}
\end{table*}

\begin{figure}[ht]
\centering
  \subfloat[][5-Class Training Dataset] {
    \centering
    \includegraphics[width=0.20\textwidth]{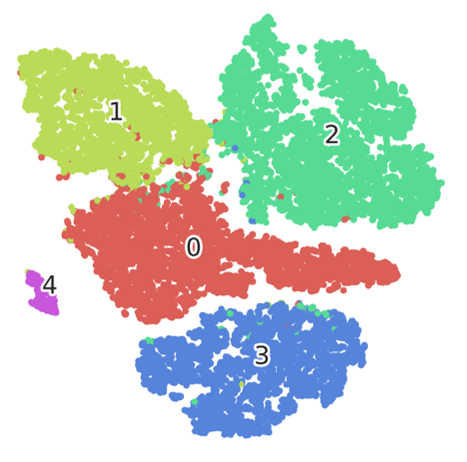}}
  \subfloat[][5-Class Test Dataset] {
    \centering
    \includegraphics[width=0.20\textwidth]{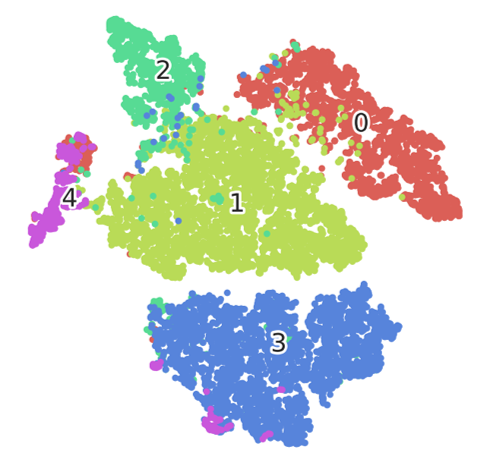}}
  \caption{t-SNE plots for 5-class model.}
  \label{fig:tsne_3}
\end{figure}

The water column patches exhibit different visual features to the predominantly substrate patches.  This led us to separate the Background class into sub-classes for Substrate and Water.  %
After retraining on this modified dataset the classifier achieved an overall accuracy of 92.4\% (Table \ref{table:with_water}), and the resulting t-SNE plots have five distinct clusters (Figure \ref{fig:tsne_3}). 

\begin{table*}[h]
\centering
\newcolumntype{a}{>{\hsize=0.7\hsize}X}
\newcolumntype{b}{>{\hsize=3.7\hsize}X}
\caption{5-Class Classifier Per-Class Precision and Recall}
\label{table:with_water}
\begin{tabularx}{\textwidth}{baaaaaaaaaaa}
\toprule
{} &
\multicolumn{2}{c}{\textbf{Strappy}} &
\multicolumn{2}{c}{\textbf{Ferny}} & 
\multicolumn{2}{c}{\textbf{Rounded}} & 
\multicolumn{2}{c}{\textbf{Substrate}} & 
\multicolumn{2}{c}{\textbf{Water}} & {}\\
\midrule
{} & \text{Prec.} & \text{Recall} & \text{Prec.} & \text{Recall} & \text{Prec.} & \text{Recall} & \text{Prec.} & \text{Recall} & \text{Prec.} & \text{Recall} & \text{Accuracy} \\
\midrule
Validation Results & 0.993 & 0.987 & 0.990 & 0.985 & 1.000 & 0.985 & 0.981 & 0.981 & 1.000 & 1.000 & 0.989 \\
Test Results & 0.892 & 0.952 & 0.938 & 0.960 & 0.898 & 0.858 & 0.94 & 0.982 & 0.982 & 0.258 & 0.924 \\
\bottomrule
\end{tabularx}
\end{table*}

\subsection{Inferences on Larger Images}

\begin{figure}[t]
\centering
  \subfloat[][Strappy] {
    \centering
    \includegraphics[width=0.20\textwidth]{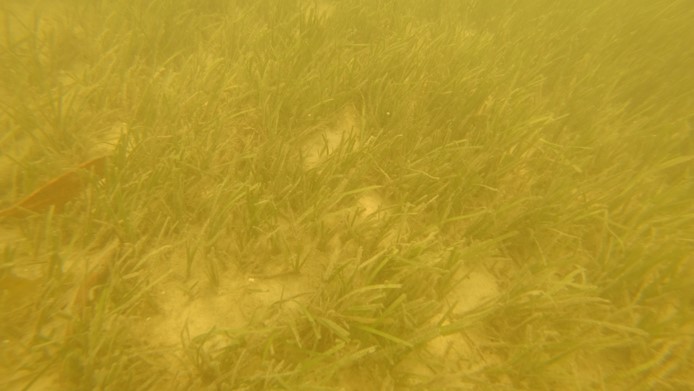}}
  \subfloat[][Ferny] {
    \centering
    \includegraphics[width=0.20\textwidth]{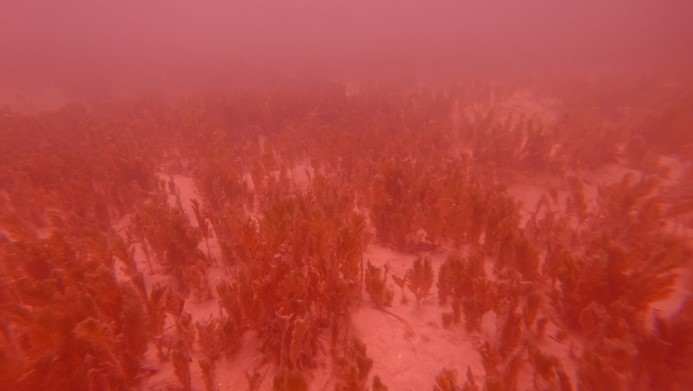}}\\
    \subfloat[][Rounded] {
    \centering
    \includegraphics[width=0.20\textwidth]{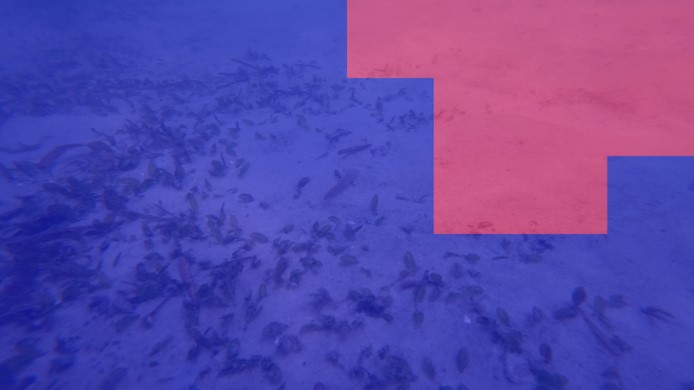}}
    \subfloat[][Background] {
    \centering
    \includegraphics[width=0.20\textwidth]{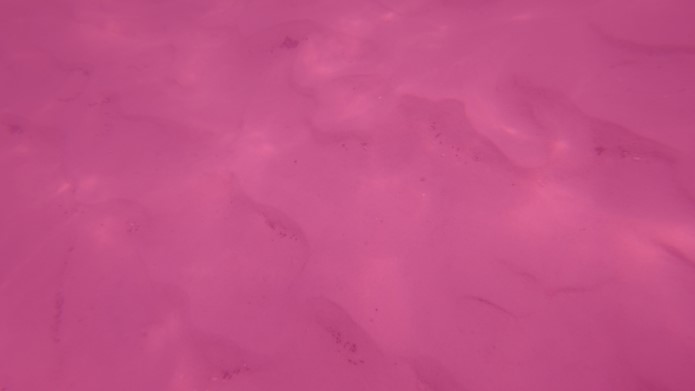}}
  \caption{Test dataset images with with correctly classified patches (where yellow = strappy, red = ferny, blue = rounded and pink = background).}
  \label{fig:good_images}
\end{figure}

The trained classifier was then applied to entire images to analyse where it had failed. Patches are classified correctly in most of the images (Figure \ref{fig:good_images}). Interestingly, final classifications are correct in some of the images where patches differ from image-level labels, as occurs frequently in less dense `Rounded' examples. This suggests that the model may classify grid cells correctly in cases where labels are ambiguous or incorrect.

\begin{figure}[t]
\centering
  \subfloat[][Strappy] {
    \centering
    \includegraphics[width=0.20\textwidth]{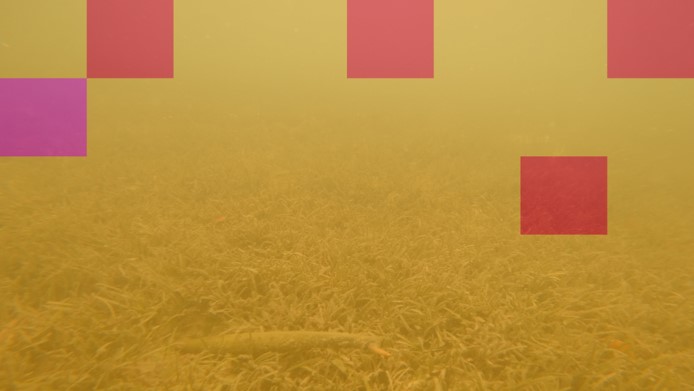}}
  \subfloat[][Ferny] {
    \centering
    \includegraphics[width=0.20\textwidth]{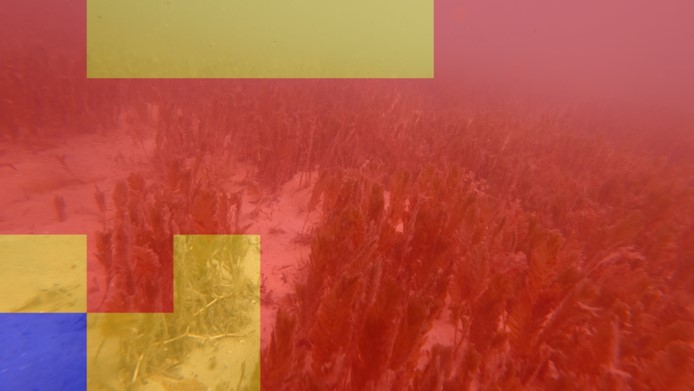}}\\
    \subfloat[][Rounded] {
    \centering
    \includegraphics[width=0.20\textwidth]{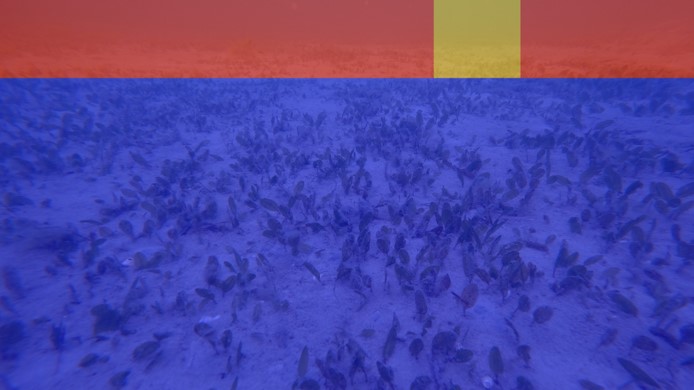}}
    \subfloat[][Background] {
    \centering
    \includegraphics[width=0.20\textwidth]{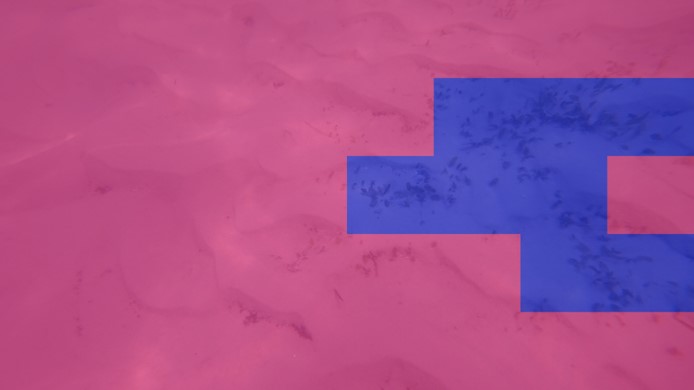}}
  \caption{Test dataset images with incorrectly classified patches \\ (where yellow = strappy, red = ferny, blue = rounded and pink = background).}
  \label{fig:bad_images}
\end{figure}

The failure case examples (Figure \ref{fig:bad_images}) suggest that blur and noise in the top-row of patches caused errors. For example, in the top row of the `Ferny' and `Rounded' examples the seafloor recedes into the distance and seagrass is too indistinct for useful classification.  These cells were deliberately removed from the training set. 
Additionally, for the `Ferny' image, the model has mistaken the exposed seagrass root system as a `strap-like' feature, triggering incorrect Strappy classifications in the bottom left corner of the frame (Figure \ref{fig:bad_images} (b)).

\section{Discussion}
\label{sec:discussion}

The method presented in this work relies on the collection of images that contain a single class.  Labelling is then only required at the image level, and labels can be propagated to sub-image patches with high confidence. 
Most of the labelling work is thus shifted to collection time, where the photographer aims to take as many single-class images as possible. Patterns of species dominance within seagrass beds make them an ideal application of the technique, but it could also be applied to other ecological applications such as detecting and classifying corals in underwater images or agricultural applications such as detecting crop diseases or weeds \cite{moghadam2017plant}.

Using this approach we were able to discriminate seagrasses from substrate and classify them into three morphotype super-classes. When visualised on whole images the patches were generally classified correctly (Figure \ref{fig:good_images}), even for patches differing in class from their image label. This suggests assigning image-level labels to patches when creating the training set is somewhat robust to errant patches. However, we did find clear benefits to including an extra class to represent the water column or far-off background.

Further work could include extension of the method to distinguish specific species of seagrass in images.  A logical next step would be to separate thin strap-like seagrasses from thick strap-like seagrasses, for example Cymodocea serrulata and Zostera muelleri, before progressing onto more visually similiar species within the Rounded super-class. The approach would additionally benefit from extensive field trials in a wider range of conditions: depth, turbidity, time of day and geographical location.  Seagrass appearance also changes with seasons, meaning that the model could be improved by training on data collected throughout periods of growth and senescence and testing of the model across years.  We will also investigate the use of domain randomized synthetic dataset to bridge domain and species gap \cite{ward2018deep}, \cite{ward2020scalable}. %

\section{Conclusion}
\label{sec:conclusions}
This work has implemented a deep learning method for automated detection and classification of seagrass species.  The approach focused on classification of seagrass into three major morphological super-classes plus background and presented a novel method of training a multi-species seagrass classifier using a dataset of single species images.  The four-class approach achieved an overall accuracy of 88.2\%, which was then improved to 92.4\% by dividing the Background class into Substrate and Water Column sub-classes.  
The approach could be extended by separating morphological super-classes into the individual seagrass species and would benefit from increased training data and robust testing in the field.  Furthermore, the approach presented could be used as a basis for estimations of species-specific percentage cover and density. To foster further research, we release the dataset used in this paper and the code developed for analysis.

\section*{Acknowledgment}
This work was done in collaboration between CSIRO Data61, CSIRO Oceans and Atmosphere, Babel-sbf and QUT and was funded by CSIRO’s Active Integrated Matter and Machine Learning and Artificial Intelligence (MLAI) Future Science Platform.  S.R., R.M. and F.M. acknowledge continued support from the Queensland University of Technology (QUT) through the Centre for Robotics.

\bibliographystyle{IEEEtran} 
\bibliography{Bibliography} 

\end{document}